\documentclass[preprint,12pt]{elsarticle}

\usepackage{amssymb}

\usepackage[utf8]{inputenc} 
\usepackage[T1]{fontenc}    
\usepackage{caption}
\usepackage{hyperref}       
\usepackage{url}            
\usepackage{booktabs}       
\usepackage{amsfonts}       
\usepackage{nicefrac}       
\usepackage{microtype}      
\usepackage{xcolor}         
\usepackage{tabularx}

\usepackage{amsmath}
\usepackage{amssymb}
\usepackage{tikz}
\usetikzlibrary{shapes.geometric, arrows}
\usepackage{booktabs}
\usepackage{adjustbox}
\usepackage{float}

\usepackage{graphicx}
\usepackage{subcaption}

\usepackage{array}
\usepackage{makecell}
\usepackage{algorithm}
\usepackage{algpseudocode}
\usepackage{booktabs}
\usepackage{hyperref} 
\usepackage{natbib}

\usepackage{import}
\usepackage{tikz}
\usetikzlibrary{positioning, calc, fit}
\usetikzlibrary{quotes,arrows.meta}
\usetikzlibrary{positioning}

\def\edgecolor{rgb:blue,4;red,1;green,4;black,3}
\newcommand{\midarrow}{\tikz \draw[-Stealth,line width =0.8mm,draw=\edgecolor] (-0.3,0) -- ++(0.3,0);}

\usepackage{Ball}
\usepackage{Box}
\usepackage{RightBandedBox}

\newcolumntype{C}{>{\centering\arraybackslash}X}
\newcolumntype{R}{>{\raggedleft\arraybackslash}X}

\newcommand{\methodfull}{Locally Competitive Algorithm with State Warm-up via Predictive Priming}
\newcommand{\method}{WARP-LCA} 

\journal{Neurocomputing}

\begin{document}

\begin{frontmatter}

\title{\method: Efficient Convolutional Sparse Coding with Locally Competitive Algorithm}

\author[inst1,inst2]{Geoffrey Kasenbacher}
\author[inst1]{Felix Ehret}
\author[inst1]{Gerrit Ecke}
\author[inst2]{Sebastian Otte}

\affiliation[inst1]{organization={Mercedes-Benz AG},
            addressline={Leibnizstraße 2}, 
            city={Böblingen},
            country={Germany}}

\affiliation[inst2]{organization={Institut für Robotik und Kognitive Systeme, Universität zu Lübeck},
            addressline={Ratzeburger Allee 160}, 
            city={Lübeck},
            country={Germany}}

\begin{abstract}
The locally competitive algorithm (LCA) can solve sparse coding problems across a wide range of use cases. Recently, convolution-based LCA approaches have been shown to be highly effective for enhancing robustness for image recognition tasks in vision pipelines. To additionally maximize representational sparsity, LCA with hard-thresholding can be applied. While this combination often yields very good solutions satisfying an \( \ell_0 \) sparsity criterion, it comes with significant drawbacks for practical application: (i) LCA is very inefficient, typically requiring hundreds of optimization cycles for convergence; (ii) the use of a hard-thresholding results in a non-convex loss function, which might lead to suboptimal minima. To address these issues, we propose the \methodfull{} (\method{}), which leverages a predictor network to provide a suitable initial guess of the LCA state based on the current input. Our approach significantly improves both convergence speed and the quality of solutions, while maintaining and even enhancing the overall strengths of LCA. We demonstrate that \method{} converges faster by orders of magnitude and reaches better minima compared to conventional LCA. Moreover, the learned representations are more sparse and exhibit superior properties in terms of reconstruction and denoising quality as well as robustness when applied in deep recognition pipelines. Furthermore, we apply \method{} to image denoising tasks, showcasing its robustness and practical effectiveness. Our findings confirm that the naive use of LCA with hard-thresholding results in suboptimal minima, whereas initializing LCA with a predictive guess results in much better outcomes.
\end{abstract}

\begin{graphicalabstract}

\begin{figure}[H]
    \centering
    \includegraphics[width=\textwidth]{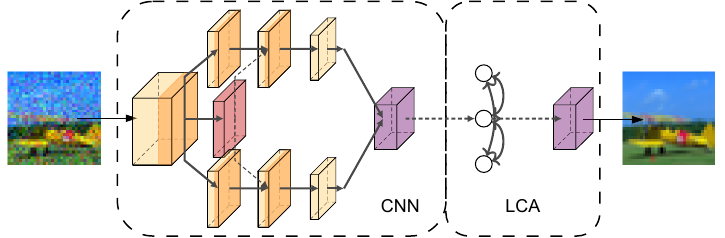}
    \caption{
    \textbf{Illustration of the WARP-LCA Method.}
    The WARP-LCA method integrates a fully convolutional neural network (CNN) to predict LCA states. 
    These predicted states serve as a warm start for the LCA module. 
    After several LCA iterations, the refined sparse activations are optionally processed through a transpose convolution block to reconstruct the image. 
    }
\end{figure}

\end{graphicalabstract}

\begin{highlights}
\item WARP-LCA accelerates convergence and achieves superior sparsity compared to LCA.
\item Achieves higher PSNR and SSIM with fewer iterations than traditional LCA.
\item Improves denoising in classification pipelines under varying noise levels.
\item Enables generalizable and efficient sparse coding with predictive initialization.
\end{highlights}


\end{frontmatter}


\section{Introduction}
\label{sec:introduction}
Sparse coding has deep roots in neuroscience. The idea that sensory systems learn and represent the statistics of natural scenes date back to Barlow's efficient coding hypothesis \citep{barlow1961possible}. Barlow reasoned that redundancy reduction is key, since information retrieval in the sensory stream is akin to finding the needle in the haystack. Sparse coding then originated from the insight that sparse, redundant representations may be useful to make the statistical components of sensory information explicit \citep{field1994what, barlow2001redundancy}. Within this context, the sparse coding algorithm, introduced by \citep{olshausen1996emergence}, has emerged as a paradigm to successfully model and predict response patterns of the primary visual cortex \citep{lee2006efficient,lee2007sparse,beyeler2019neural,ecke2021exploitation}. 

The human cortex is a very homogeneous structure, and it has been hypothesized that it performs the same operation everywhere \citep{barlow1987cerebral}. The claim that sparse coding, at least to some degree, models general cortical function, reflects in a manifold of applications in which the algorithm has proven useful. It has early on been shown that the learned features are useful for inference \citep{rigamonti2011are}, especially when unsupervised learning can be leveraged, like in stereo vision \citep{lundquist2016sparse,lundquist2017sparse,ecke2021exploitation} or in scenarios with poorly labeled data \citep{zhang2017medical}. The capabilities for unsupervised learning are most prominent in multi layer sparse coding networks, which are capable of learning abstract concepts \citep{kim2018deep,zhang2019hierarchical,dibbo2023LCANets}. In addition, sparse coding has proven at least state-of-the-art performance in removing noise and other corruptions from images \citep{lecouat2020fully}. More recently, \citet{teti2022lcanets} have shown that applying sparse coding as a pre-processing stage in image recognition pipelines yields state-of-the-art robustness to common corruptions and adversarial attacks. 

A significant challenge for applications with sparse coding is the computational expense of the inference algorithm. Finding sparse coefficients for each input sample requires an expensive iterative optimization process via gradient descent. One approach to resolve poor execution speed is the implementation on neuromorphic hardware. \citet{rozell2008sparse} introduced the Locally Competitive Algorithm (LCA) as a biologically plausible model for sparse coding. 
Indeed, the implementation of the LCA on Intel Loihi has proven the strongest gain in energy efficiency and execution speed against the implementation on conventional hardware, as compared to any other implemented algorithm \citep{davies2021advancing,henke2022apples,parpart2023implementing}. Considerable research has been directed towards developing efficient optimization algorithms for sparse coding on conventional hardware. Notable contributions include the works of \citet{li2009compressed,mairal2009online,beck2009fast}, and \citet{vonesch2007generalized}. \citet{gregor2010learning} contributed significantly to the field by introducing the Learned ISTA (LISTA) algorithm, which has spawned a family of model-based sparse encoders (notable models include AMP-Net \citet{zhang2020amp}, FISTA-Net \citet{xiang2021fista} and ISTA-Net \citet{zhang2018ista}. 

Sparse coding is typically associated with \( \ell_1 \) norm regularization due to its convexity, which facilitates a tractable optimization process \citep{zhang2015survey}. However, for achieving maximum sparsity, \( \ell_0 \)-like regularization is preferred \citep{paiton2019analysis}. Despite its non-convex nature, which introduces computational challenges, \( \ell_0 \) regularization directly targets sparsity in the solution \citep{nguyen2019np}. Nonetheless, the Locally Competitive Algorithm (LCA) presents a distinctive case. While the discrete version of LCA implementing \( \ell_1 \) sparsity aligns mathematically with ISTA \ \citep{balavoine2015discrete}, employing an \( \ell_0 \)-like cost function in LCA diverges significantly in outcomes when compared to greedy methods such as Basis Pursuit (BP) \citep{rozell2007locally,rozell2008sparse}.

Our main contribution of this paper is the development and validation of the Locally Competitive Algorithm with State Warm-up via Predictive Priming (WARP-LCA). This novel approach leverages a predictor network to provide an initial guess of the LCA state based on the current input, significantly improving convergence speed and solution quality. WARP-LCA addresses the inefficiencies and suboptimal solutions associated with LCA using hard-thresholding (\( \ell_0 \)). By incorporating predictive priming, WARP-LCA dramatically reduces the number of optimization cycles required for convergence. Furthermore, WARP-LCA achieves superior minima, yielding more sparse and robust representations that enhance reconstruction and denoising quality, particularly in deep recognition pipelines. Our experiments, in the domain of image denoising and comparing WARP-LCA directly to LCA, confirm that WARP-LCA not only overcomes the limitations of traditional hard-thresholding LCA but also advances the field of biologically inspired deep learning by providing a novel and efficient method for convolutional sparse coding.

The remainder of this paper is structured as follows: Section 2 provides the background and related work, elaborating on convolutional sparse coding and the specifics of the Locally Competitive Algorithm. Section 3 details the WARP-LCA method, particularly focusing on the integration of the predictor network. In Section 4, we present our experimental setup and results, demonstrating the efficacy of WARP-LCA over traditional LCA in scenarios of image denoising and classification robustness. Section 5 discusses the broader implications of our findings and explores the generalizability of WARP-LCA. Finally, Section 6 concludes the paper by summarizing our contributions and suggesting ways in which WARP-LCA could be applied to further applications and extended in future work.

\section{Background and Related Work}

\subsection{Convolutional Sparse Coding}

Convolutional sparse coding (CSC) is a technique used in signal processing and machine learning to learn a representation of data through the joint processes of inference and learning. The core idea is to represent input data using a sparse combination of basis functions.

Let the input image \( x \in \mathbb{R}^{C \times H \times W} \) of height \( H \) and width  \( W \), with \( C \) channels, be represented by a learned overcomplete dictionary of convolutional features \( \Phi \in \mathbb{R}^{M \times C \times k_H \times k_W} \), where \( M \) is the number of convolutional features, and \( k_H \) and \( k_W \) represent the spatial dimensions of each kernel. The input \( x \) can be represented as \( x = \Phi \circledast a \), where \( \circledast \) denotes the transposed convolution operation and \( a \in \mathbb{R}^{M \times \left\lfloor \frac{H}{\text{stride}_H} \right\rfloor \times \left\lfloor \frac{W}{\text{stride}_W} \right\rfloor} \) is the coefficient vector.

The inference process involves estimating the sparse coefficients \( a \) for a given data point \( x \). This is typically achieved by minimizing an energy function that combines a reconstruction error term with a sparsity-inducing regularization term. A common formulation uses \( \ell_p \) regularization to enforce sparsity \citep{tibshirani1996regression}.
\begin{equation}
E(x, \Phi, a) = \|x - \Phi \circledast a\|_2^2 + \lambda \|a\|_p
\end{equation}

Here, \(\lambda\) controls the strength of the sparsity constraint, balancing the trade-off between the reconstruction error and the sparsity of the coefficients.  The choice of the sparsity criterion (i.e., the norm used for \( a \)) significantly affects the magnitude of the coefficients and the overall sparsity of the representation.

Learning in convolutional sparse coding involves updating the convolutional filters \(\Phi\) based on the inferred sparse coefficients across a batch of data points. This is typically done through gradient descent on the energy function with respect to \(\Phi\). Although the joint optimization of \(\Phi\) and \(a\) is non-convex, iterative approaches often converge to useful solutions, especially in natural signal domains. In addition to convolutional sparse coding, other widely known approaches for learning dictionaries include Principal Component Analysis (PCA) and Independent Component Analysis (ICA). PCA learns an orthogonal dictionary that captures the directions of maximum variance in the data. ICA, on the other hand, seeks to represent data as linear combinations of statistically independent components, which is useful for blind source separation and feature extraction \citep{Hyvarinen2000}.

\subsection{Locally Competitive Algorithm (LCA)}

The Locally Competitive Algorithm (LCA), introduced by Rozell et al. \cite{rozell2008sparse}, represents a biologically plausible approach to sparse coding, where the dynamics of the neuronal population can be conceptualized as a system of coupled differential equations that govern the temporal evolution of neuronal activities. Unlike iterative thresholding methods or matching pursuit, the LCA exhibits a 'charging circuit behavior' where neurons dynamically compete to represent the input.

To more intuitively describe the governing dynamics of the neurons in the context of Convolutional LCA, the states \(u_i\) and activations \(a_i\) refer to a map of neurons (neural map \(i\)), with the dimension \( \frac{H}{\text{stride}_H} \) and \( \frac{W}{\text{stride}_W} \), that correspond to a single kernel \(\phi_i\). 

The dynamics of each neural map's membrane potential $u_i$ follows the differential equation

\begin{equation}
\tau \frac{du_i}{dt} = -u_i + \phi_i \ast x + a_i - \sum_{i \neq j} a_j \ast (\phi_i \ast \phi_j)  \, , %
\end{equation}
Activations \(a_i\) are derived from a thresholding function applied to \(u_i\):

\begin{equation}
a_i = T_\lambda(u_i) \, .
\end{equation}

% ...
where \(\tau\) is a time constant that controls the rate of membrane potential decay. The self-inhibition term $-u_i$ dampens the dynamics, and the term $\phi_i \ast x$ represents the feedforward input to neural map i. The \(+ a_i\) eliminates self interaction from the following term. $\sum_{i \neq j} a_j \ast (\phi_i \ast \phi_j)$ couples the differential equations and can be seen as lateral competition between neural maps. 

\citet{rozell2008sparse} introduced a generalized thresholding function that enables the Locally Competitive Algorithm (LCA) to tackle various sparse coding problems by adjusting its parameters. The generalized thresholding function is expressed as:

\begin{equation}
T_\lambda(u_i) = \frac{u_i - \alpha \lambda}{1 + e^{-\gamma (u_i - \lambda)}}
\end{equation}

In this equation, \( \lambda \) is a parameter that controls the sparsity level, \( \alpha \) scales the threshold, and \( \gamma \) adjusts the steepness of the threshold function. By fine-tuning these parameters, the LCA can transition from soft-thresholding (promoting \( \ell_1 \) sparsity) to hard-thresholding (promoting \( \ell_0 \) sparsity).

Specifically, for \( \ell_0 \) regularization—which aims to minimize the number of non-zero coefficients—the thresholding function approaches a hard threshold as \( \gamma \) becomes large. This effectively turns the generalized thresholding function into the well-known proximal operator:

\begin{equation}
T_\lambda(u_i) =
\begin{cases}
u_i - \alpha \lambda, & \text{if } u_i > \lambda \\
0, & \text{otherwise}
\end{cases}
\end{equation}

This hard-thresholding function enforces strict sparsity by zeroing out coefficients below the threshold \( \lambda \), aligning with the goals of \( \ell_0 \) regularization.

The initial values of the states in the LCA are typically set to zeros \citep{rozell2007locally}, which can lead to slower convergence, especially under \( \ell_0 \) constraints due to the non-convexity of the optimization landscape. This issue underscores the importance of effective initialization strategies to accelerate convergence when employing \( \ell_0 \) regularization in LCA.

The LCA framework has demonstrated significant versatility in solving a range of sparse coding problems by appropriately setting the thresholding parameters. This adaptability makes the LCA a robust tool for sparse representation learning in diverse applications. Its major drawback, however, is that it becomes computationally intensive and slow when executed on standard computing hardware, particularly with \( \ell_0 \) regularization, necessitating optimization techniques to improve its efficiency.

Recent implementations of the LCA on neuromorphic hardware have alleviated this problem to a certain extent by significantly enhancing the computational efficiency and execution speed by several orders of magnitude \citep{davies2021advancing,henke2022apples,parpart2023implementing}. Despite this progress, the algorithm still requires hundreds of optimization iterations and encounters substantial challenges with local optima. Thus, even with neuromorphic hardware, there is significant room for improvement in the underlying algorithmic principles.

\subsection{Warm Starting with Deep Learning}

A common approach to enhance the efficiency of iterative algorithms is to learn a mapping from problem parameters to high-quality initializations, known as warm starting. This technique leverages deep learning to predict an initial state that accelerates the convergence of iterative solvers.

\citet{sambharya2023learning} introduced a framework for learning warm starts specifically for Douglas-Rachford splitting to solve convex quadratic programs (QPs), providing generalization guarantees for a broad range of operators. Similarly, \citet{klauvco2019machine} used neural networks to warm start active set methods in model predictive control (MPC), demonstrating significant reductions in computation time.

Beyond convex optimization, deep learning has been applied to accelerate non-convex optimization algorithms. For example, \citet{sjolund2022graph} used graph neural networks to expedite matrix factorization, and developed schemes for quickly solving fixed-point problems. Inverse problems, such as sparse coding (\citep{gregor2010learning}, \citep{xin2016maximal}), image restoration \citep{rick2017one}, and wireless communication \citep{he2020model}, have also benefited from embedding algorithm steps into deep networks. A widely used technique is unrolling algorithmic steps, which differentiates through these steps to minimize a performance loss \citep{monga2021algorithm, diamond2017unrolled, gregor2010learning}.

\section{WARP-LCA}
Our approach introduces the \textbf{\methodfull{} (\method{})}, which integrates a predictor network into the traditional LCA framework. More specifically we propose a fully convolutional network for this purpose. The key contribution of our approach lies in the CNN's role in predicting the initial states $u_i$ for the LCA, facilitating faster convergence and enhancing the quality of the sparse code. \autoref{fig:warp_lca} shows a simple illustration of the approach.

The network is trained on internal states obtained from running the traditional LCA for many iterations and optionally for different sparsity levels. The sparsity level is encoded into the input data as an additional constant channel (i.e. RGB$\lambda$).

\begin{figure}[t!]
  \centering
  \colorlet{ConvColor}{rgb:yellow,5;red,2.5;white,5}
\colorlet{ConvReluColor}{rgb:yellow,5;red,5;white,5}
\colorlet{PoolColor}{rgb:red,1;black,0.3}
\colorlet{DcnvColor}{rgb:blue,5;green,2.5;white,5}
\colorlet{SoftmaxColor}{rgb:magenta,5;black,7}
\colorlet{EdgeColor}{rgb:blue,2;red,2;green,2;black,1}

\begin{tikzpicture}[
    scale=0.75,
    transform shape,
    font=\footnotesize
]
\tikzstyle{connection}=[ultra thick, every node/.style={sloped,allow upside down},
                        draw=EdgeColor,opacity=0.7]
\tikzstyle{skipconn}=[densely dashed,thick,EdgeColor, every node/.style={sloped,allow upside down}]

\pic[shift={(0,0,0)}]
  at (0,0,0)
  {RightBandedBox={
    name=block1and2,
    caption={Block: conv1–6},
    xlabel={{"512",""}},
    fill=ConvColor,
    bandfill=ConvReluColor,
    height=12,
    width=7,
    depth=12
  }};

\pic[shift={(1.4,0,0)}]
  at (block1and2-east)
  {Box={
    name=adj,
    caption={},          
    xlabel={{"256",""}},
    fill=PoolColor,
    height=10,
    width=2,
    depth=10
  }};

\pic[shift={(1.4,-3.,0)}] 
  at (block1and2-east)
  {RightBandedBox={
    name=d1,
    caption={Branch 1: conv1},
    xlabel={{"256",""}},
    fill=ConvColor,
    bandfill=ConvReluColor,
    height=10,
    width=2,
    depth=10
  }};

\pic[shift={(1.4,0,0)}]
  at (d1-east)
  {RightBandedBox={
    name=d2,
    caption={Branch 1: conv2},
    xlabel={{"256",""}},
    fill=ConvColor,
    bandfill=ConvReluColor,
    height=10,
    width=2,
    depth=10
  }};

\pic[shift={(1.4,0,0)}]
  at (d2-east)
  {Box={
    name=d3,
    caption={Branch1: conv3},
    xlabel={{"100",""}},
    fill=ConvColor,
    height=8,
    width=1.5,
    depth=8
  }};

\pic[shift={(1.4 ,3. ,0)}]
  at (block1and2-east)
  {RightBandedBox={
    name=u1,
    caption={},
    xlabel={{"256",""}},
    fill=ConvColor,
    bandfill=ConvReluColor,
    height=10,
    width=2,
    depth=10
}};

\pic[shift={(1.4,0,0)}]
  at (u1-east)
  {RightBandedBox={
    name=u2,
    caption={},
    xlabel={{"256",""}},
    fill=ConvColor,
    bandfill=ConvReluColor,
    height=10,
    width=2,
    depth=10
  }};

\pic[shift={(1.4,0,0)}]
  at (u2-east)
  {Box={
    name=u3,
    caption={Branch 2: conv3},
    xlabel={{"100",""}},
    fill=ConvColor,
    height=8,
    width=1.5,
    depth=8
  }};

\pic[shift={(2.0,0,0)}]
  at ($(d3-east)!0.5!(u3-east)$)
  {Box={
    name=out,
    caption={Output},
    xlabel={{"100",""}},
    fill=SoftmaxColor,
    height=8,
    width=4,
    depth=8
  }};

\draw [connection] (block1and2-east) -- node {\midarrow} (adj-west);

\draw [connection] (block1and2-east) -- ++(0,-1.5,0) -- node {\midarrow} (d1-west);
\draw [connection] (d1-east) -- node {\midarrow} (d2-west);
\draw [skipconn] (adj-east)  -- ++(0,-1.5,0) --  node {\midarrow} (d2-west);
\draw [connection] (d2-east) -- node {\midarrow} (d3-west);
\draw [connection] (d3-east) -- ++(0.6,0,0);

\draw [connection] (block1and2-east) -- ++(0,1.5,0) -- node {\midarrow} (u1-west);
\draw [connection] (u1-east) -- node {\midarrow} (u2-west);
\draw [skipconn] (adj-east)  -- ++(0,1.5,0) -- node {\midarrow}  (u2-west);
\draw [connection] (u2-east) -- node {\midarrow} (u3-west);
\draw [connection] (u3-east) -- ++(0.6,0,0);

\draw [connection] ($(d3-east)+(0.6,0,0)$) -- node {\midarrow} (out-west);
\draw [connection] ($(u3-east)+(0.6,0,0)$) -- node {\midarrow} (out-west);

\coordinate (lcaAnchor) at ([xshift=2.8cm] out-east);

\node[draw, thick, circle, minimum size=0.6cm, align=center] (a2)
  at (lcaAnchor) {$a_{2}$};

\node[draw, thick, circle, minimum size=0.6cm, align=center,
      above=1.0cm of a2] (a1) {$a_{1}$};

\node[draw, thick, circle, minimum size=0.6cm, align=center,
      below=1.0cm of a2] (a3) {$a_{3}$};

\draw [connection, densely dashed] (out-east) -- node {\midarrow} (a2.west);

\draw [connection, <->] (a1) to[out=340, in=20] (a2);
\draw [connection, <->] (a2) to[out=340, in=20] (a3);
\draw [connection, <->] (a1) to[out=290, in=70] (a3);

\coordinate (lcaOutAnchor) at ([xshift=2.8cm] a2);

\pic[shift={(0,0,0)}]
  at (lcaOutAnchor)
  {Box={
    name=lcaOut,
    caption={LCA Output},
    xlabel={{"100",""}},
    fill=SoftmaxColor,
    height=8,
    width=4,
    depth=8
  }};

\draw [connection, densely dashed] (a2.east) -- node {\midarrow} (lcaOut-west);

\end{tikzpicture} 
  \caption{
    \textbf{Illustration of the WARP-LCA Method.}
    The WARP-LCA method integrates a fully convolutional neural network (CNN) to predict LCA states. 
    These predicted states serve as a warm start for the LCA module. 
    After several LCA iterations, the refined sparse activations are optionally processed through a transpose convolution block to reconstruct the image. For a detailed description see \autoref{tab:branchnet}.
    }
  \label{fig:warp_lca}
\end{figure}

\subsection{Predictor Network and Integration with LCA}
The primary function of the predictor network is to provide an initial guess for the states $u_i$ of the LCA, which are then refined through iterative optimization. This approach addresses the inefficiencies inherent in traditional LCA, which often requires numerous iterations to converge. Inspired by the findings of \citet{xin2016maximal}, which demonstrate deep networks' ability to recover minimal \( \ell_0 \)-norm representations in high dimensional scenarios, our methodology strategically employs a predictive model to initialize the LCA. 

Our network predicts the states $u_i$ of the LCA rather than the activations $a_i$ directly, a decision informed by empirical observations. You may find a detailed training procedure of WARP-LCA attached in the appendix C. Predicting activations, inherently sparse, resulted in suboptimal solutions, whereas predicting states, which typically exhibit no sparsity, proved more effective. The network effectively learns to predict a 'mean' sparse code---essentially the most probable sparse representation conditioned on the input. This is conceptualized as the network learning the conditional expectation $E[S | X] \approx f(X; \theta^*)$, where $f(X; \theta)$ maps an input $X$ to an optimal state $S$, and $\theta^*$ are the optimized parameters. 

This predictive initialization significantly enhances the LCA's performance by positioning the algorithm near an (expected) optimal starting point in the solution space, as inferred from the training data. It not only accelerates convergence but also ensures that the resultant sparse representations are more robust, effectively bypassing poorer minima that the LCA alone might settle into. Thus, by leveraging the statistical regularities of sparse codes learned during training, the model not only speeds up the optimization process but also elevates the fidelity of the encoded features, contributing to both computational efficiency and enhanced solution quality.

\subsection{Designing and Training Networks to Predict LCA States}

Our goal was to identify an effective trade-off between model complexity and prediction quality without exhaustively surveying the full range of possible CNN configurations (see \autoref{tab:branchnet_forward} for more details). We investigated three main architectures across five size configurations: a simple feedforward network, a ResNet-inspired network, and a fully convolutional variant with dual branches and a subtractive arrangement in the output layer (titled WARP-CNN). Our preliminary experiments indicated that the WARP-CNN with a subtractive ReLu arrangement as described in \citet{Lang2023} balanced the joined requirements on sparsity and accurate output in WARP-CNNs:
\begin{equation}
    {u}_i = \sigma\bigl(\operatorname{relu}({a}_i) - \operatorname{relu}({b}_i)\bigr),
\end{equation}
where $a_i$ and $b_i$ are outputs of the two computation branches in the network and $\sigma$ denotes the sigmoid function. This construction establishes a stable reference that is neither the minimum nor the maximum, allowing the branches to specialize on ranges above and below this reference value, respectively.
 
Training employed the ADAM optimizer (learning rate $10^{-4}$). Smaller batch sizes consistently yielded lower losses, indicating that finer gradient updates helped avoid local minima. This arrangement reduced the loss landscape complexity and stabilized convergence. Moreover, the sparse codes used as targets were scaled to fall within $[0,1]$, to be compatible with the sigmoidal subtractive ReLU Activation as described above. Notably, the largest WARP-CNN model demonstrated superior performance among the tested setups (see \ref{tab:models_summary} and \ref{fig:warp_lca} for a full description of the model architecture). Overfitting was largely absent for larger models, whereas smaller variants exhibited plateauing validation losses, suggesting limited representational capacity rather than typical overfitting. Such plateaus are consistent with the networks converging on convolutional kernel representations for image patches, as expected when capturing the underlying sparse code structures.

{We introduce a custom loss function derived from a Laplacian prior, which can be interpreted as an approximate negative log-likelihood under a scale parameter 
\(\beta_i = 1 + \epsilon + \gamma |t_i|\). Specifically, the Laplace log-density for a target \(t_i\) given an output \(o_i\) is
\[
-\log p(t_i \mid o_i) \;\approx\; \log \bigl(\beta_i\bigr) + \frac{|t_i - o_i|}{\beta_i}.
\]
Although a Laplace prior typically implies an absolute deviation term, we adopt a squared-error surrogate for smoother gradients and more stable optimization, since the absolute-value function is not differentiable at zero. Since $\log \bigl(\beta_i\bigr)$ does not affect gradients with respect to model parameters it can be safely dropped for training objectives. Replacing the absolute deviation with a squared term yields:
\begin{equation}
\mathcal{L} 
= \frac{1}{N} \sum_{i=1}^N 
\left(
  (o_i - t_i)^2 \cdot \frac{1}{1 + \epsilon + \gamma |t_i|}
\right).
\end{equation}
Here:
\begin{itemize}
  \setlength\itemsep{-.5em}
  \item \(N\) is the total number of elements,  
  \item \(o_i\) and \(t_i\) are the model output and target,  
  \item \(\gamma\) is the Laplace scale parameter controlling how strongly to penalize deviations at small \(t_i\),  
  \item \(\epsilon\) is a small constant for numerical stability.
\end{itemize}}

{In our experiments, this custom loss effectively prioritizes small, frequent values---critical for modeling sparse targets. A value of \(\gamma = 3\) proved consistently effective. Consequently, errors at small targets are multiplied by a larger factor (due to the smaller denominator), while errors at larger targets are down-weighted. Since the target values fall within $[0,1]$, \(\gamma = 3\)  places roughly four times greater emphasis (ignoring $\epsilon$) on small $t_i$-values than on $t_i$ near one.}

\section{Experiments and Results}\label{sec:experiments}

\begin{figure}[H]
    \centering
    \includegraphics[width=0.7\textwidth]{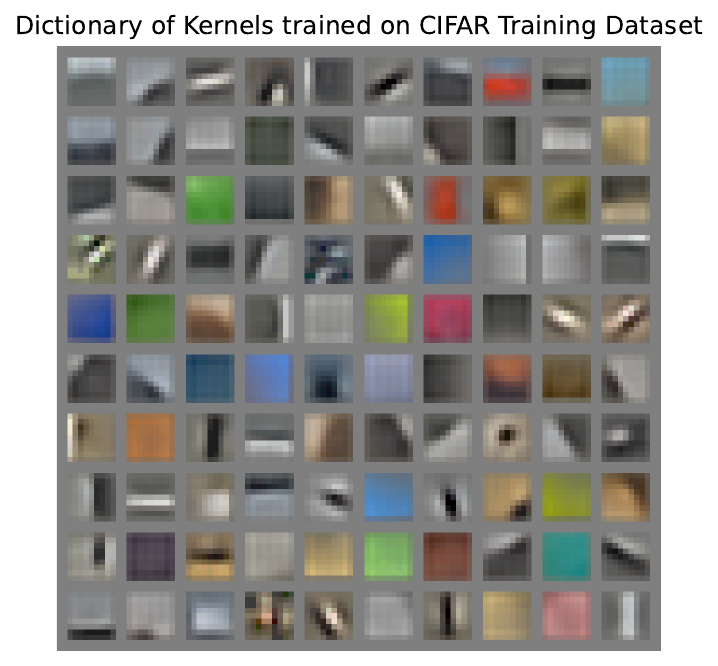}
    \caption{\\ An exemplary dictionary of kernels obtained from training on the the CIFAR dataset}
    \label{cifar_dict}
\end{figure}

In our experiments, we used the CIFAR-10 dataset \citep{krizhevsky2009learning}, the STL-10 dataset \citep{coates2011analysis}, and the Tiny ImageNet dataset \citep{tiny_imagenet} for training the models (WARP-CNN). The dictionary was learned using a soft-thresholding (you may find the dictionary in \autoref{cifar_dict}). 
For all of our experiments we normalized the input images to zero-mean and unit variance and also normalized the learned kernels of the dictionary. To train and normalize the dictionary we used the same unsupervised projected gradient based method as in \citep{teti2022lcanets}.
Additionally, to test the versatility of our method on out-of-distribution data, we employed the Oxford Pets dataset \citep{parkhi2012cats} and a high-resolution image from Pexels \citep{pexels_image}. For all of our experiments we used the convolutional LCA LCA package \footnote{\url{https://github.com/lanl/lca-pytorch/tree/main}} presented by \citet{teti2022lcanets}.
The experiments were conducted on a Linux system equipped with three NVIDIA Tesla V100-PCIE-16GB GPUs, each with a compute capability of 7.0 and 16.94 GB of memory. The system also features 8 CPU cores and memory resources, with a total of 201.18 GB of RAM and 174.20 GB available. 

The primary task and comparison for our WARP-LCA model was against the traditional LCA method. This choice allowed for a direct evaluation of the enhancements offered by our predictive initialization approach under identical encoding conditions. 

Performance was assessed using Mean Squared Error (MSE), \( \ell_0 \)-norm, Peak Signal-to-Noise Ratio (PSNR), and Structural Similarity Index (SSIM). These metrics were chosen to provide a comprehensive view of both quantitative and qualitative measures. Each model was executed for 1\,000 iterations with $\lambda$ set to 0.15 and $\tau$ at 200.

\begin{figure}[t!]
    \centering    \includegraphics[width=\textwidth]{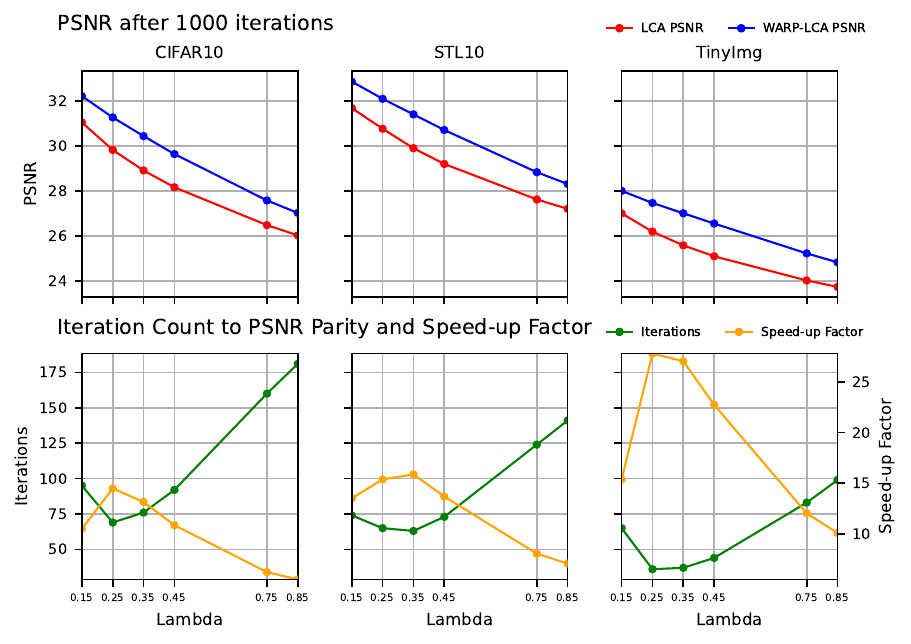}
    \caption{\textbf{Comparative Speed-Up of WARP-LCA Relative to LCA.} {The top panels show the maximum PSNR obtained after 1,000 iterations by LCA (red) and WARP-LCA (blue) for CIFAR-10, STL-10, and Tiny ImageNet, illustrating that the PSNR values for both methods typically decrease as the sparsity parameter (i.e., $\lambda$) increases. The bottom panels present the iteration count at which WARP-LCA equals LCA’s PSNR benchmark, alongside the speed-up factor (ratio of LCA’s total iteration count to WARP-LCA’s required count). The speed-up factor generally increases with larger $\lambda$, exhibiting a dataset-dependent minimum within the $\lambda \in [0.15, 0.35]$ interval, thereby highlighting WARP-LCA’s accelerated convergence to equivalent reconstruction quality.}}
    \label{fig:speed_up}    
\end{figure}

To analyze speed improvements, we evaluated models using a \(\lambda\) value of 0.15 and a $\tau$ fo 200. Speed comparisons were based on the maximum PSNR achieved by LCA after 1\,000 iterations. Since model inference allows control over the number of iterations, we compared both WARP-LCA and standard LCA by evaluating their performance after 1000 iterations to ensure full convergence. The results can be seen in \autoref{fig:speed_up}, where the WARP method achieved the highest speed up for Tiny Imagenet, where WARP-LCA matches the PSNR of LCA below 100 iterations for all chosen lambdas.

\begin{figure}[t!]
    \centering
    \includegraphics[width=1\textwidth]{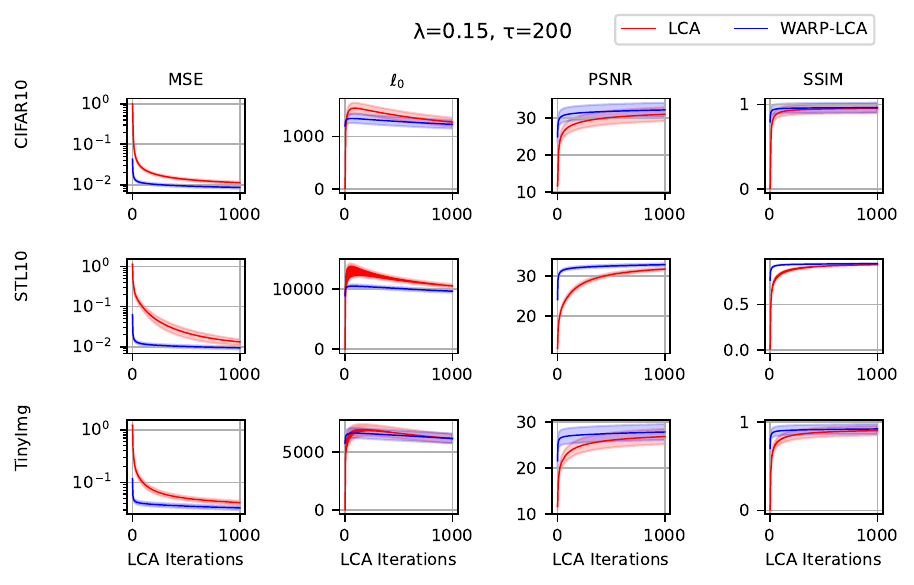}
        \caption{
    \textbf{Encoding Performance Comparison between WARP-LCA and LCA at $\lambda = 0.15$.}
    The figure displays the performance metrics for WARP-LCA and LCA. 
    The columns represent different metrics: Mean Squared Error (MSE) in log-scale, \( \ell_0 \)-Norm, Peak Signal-to-Noise Ratio (PSNR), and Structural Similarity Index (SSIM). 
    For each plot, WARP-LCA and LCA Mean and Standard Deviation of the Metrics are shown for comparison.} 
    \label{fig:compare_multiple_metrics}
\end{figure}

\begin{table}[h!]
    \centering
    \scriptsize
    \caption{Evaluation after last iteration on all datasets with $\lambda = 0.15$, $\tau = 200$}
    \begin{tabularx}{\linewidth}{lRR}
        \toprule
        Metrics & LCA Metrics (Mean) & WARP-LCA Metrics (Mean) \\
        \midrule
        \multicolumn{3}{l}{\textbf{CIFAR-10}} \\
        \midrule
        Recon. MSE & 0.0112 & \bf{0.0086} \\
        $L_0$ Norm & 1\,274.97 & \bf{1\,232.43} \\
        PSNR       & 30.98   & \bf{32.15} \\
        SSIM       & 0.95    & \bf{0.96} \\
        \midrule
        \multicolumn{3}{l}{\textbf{STL-10}} \\
        \midrule
        Recon. MSE & 0.0133 & \bf{0.0095} \\
        $L_0$ Norm & 10\,522.21 & \bf{9\,668.79} \\
        PSNR       & 31.68   & \bf{32.79} \\
        SSIM       & 0.94    & \bf{0.95} \\
        \midrule
        \multicolumn{3}{l}{\textbf{Tiny Imagenet}} \\
        \midrule
        Recon. MSE & 0.0413 & \bf{0.0328} \\
        $L_0$ Norm & \bf{6\,122.32} & 6\,158.49 \\
        PSNR       & 26.82   & \bf{27.81} \\
        SSIM       & 0.90    & \bf{0.92} \\
        \bottomrule
    \end{tabularx}
    \label{tab:table_results}
\end{table}

Our results, presented in \autoref{fig:compare_multiple_metrics} and in \autoref{tab:table_results}, showcase the benefits of using WARP-LCA. The WARP-LCA demonstrated markedly faster convergence in terms of MSE, as indicated by the log(MSE) trajectory, maintaining a superior MSE even after 1,000 iterations compared to the standard LCA. Similarly, \( \ell_0 \)-norm results showed that WARP-LCA began near optimal sparsity levels, whereas LCA required considerable iterations to approximate these levels. Notably, the \( \ell_0 \)-norm achieved by WARP-LCA was lower than that reached by LCA, indicating more efficient sparse coding. Both PSNR and SSIM metrics reflected faster convergence and higher final values for WARP-LCA across all datasets. 

This trend was consistent across all datasets, with the only exception for Tiny ImageNet, where the final average sparsity (\( \ell_0 \)-norm) was slightly lower for standard LCA.

\subsection{Qualitative Comparison of WARP-CNN Outputs and One-Step LCA Refinement}
\begin{figure}[t!]
    \centering
    \begin{subfigure}[t]{0.48\textwidth}
        \centering
        \includegraphics[width=\textwidth]{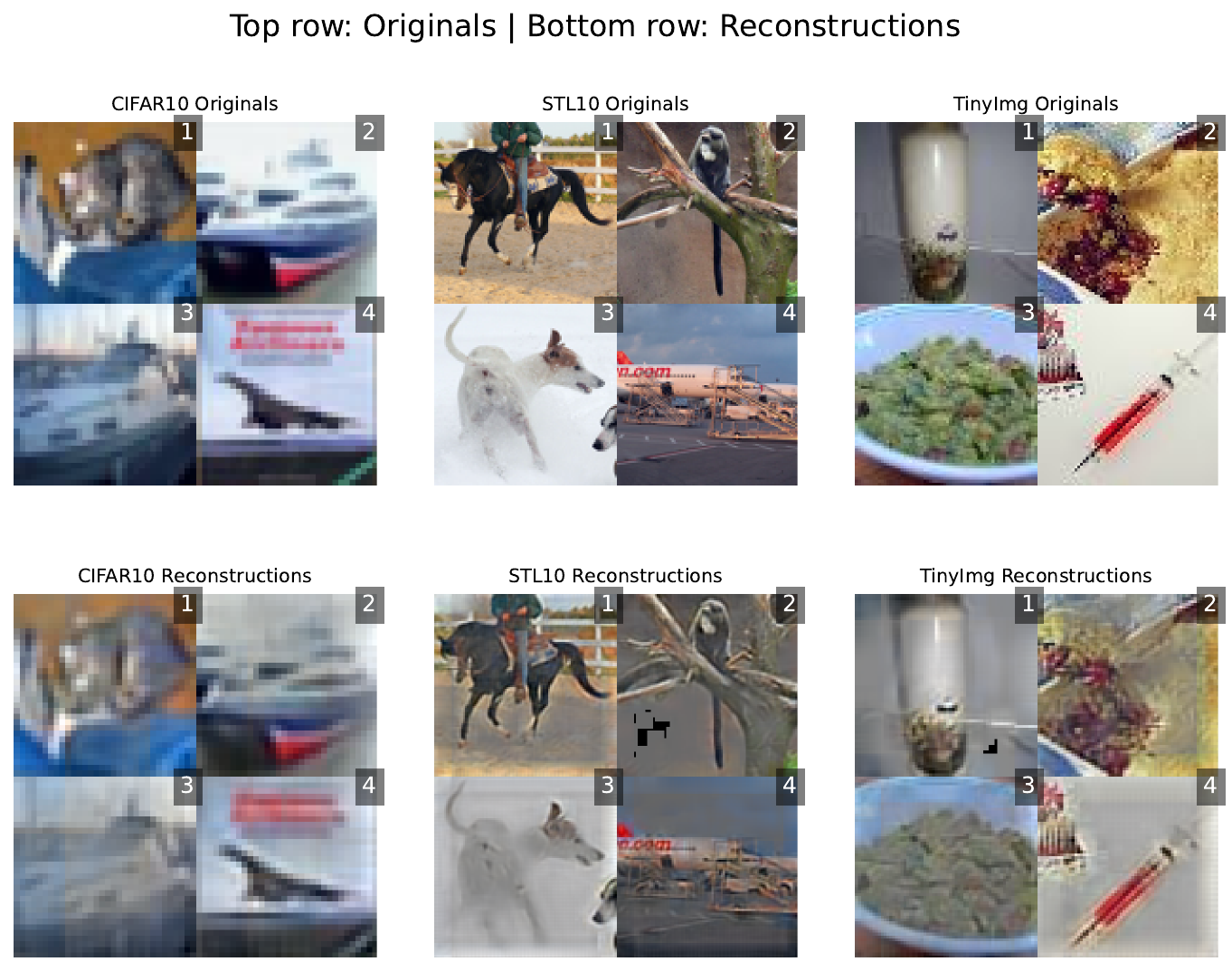}
        \caption{\textbf{WARP-CNN Without LCA.} The figure shows images of CIFAR-10, STL-10, and Tiny ImageNet along with the reconstructions obtained from the respective trained WARP-CNN models.}
        \label{fig:collage1} 
    \end{subfigure}
    \hfill
    \begin{subfigure}[t]{0.48\textwidth}
        \centering
        \includegraphics[width=\textwidth]{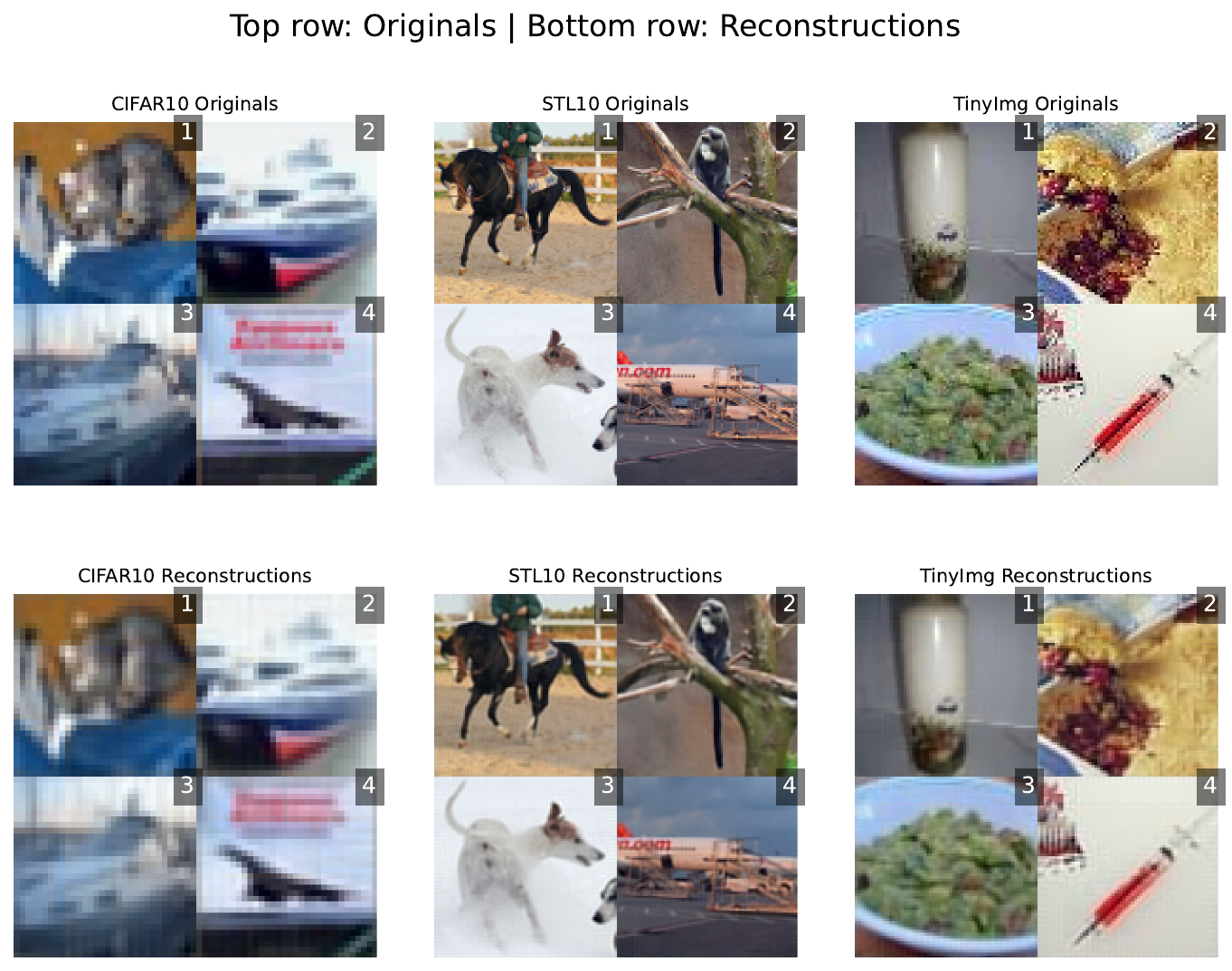}
        \caption{\textbf{WARP-CNN With One LCA Iteration.} The figure shows images of CIFAR-10, STL-10, and Tiny ImageNet and the reconstructions from a single LCA iteration with a state prediction obtained from the respective trained WARP-CNN models. The selected \(\tau\) (learning rate) was set to 70.}
        \label{fig:collage2}
    \end{subfigure}
    
    \caption{\textbf{Side-by-Side Comparison of WARP-CNN Reconstructions.} Comparison of WARP-CNN outputs and WARP-CNN with one LCA iteration across CIFAR-10, STL-10, and Tiny ImageNet datasets.}
    \label{fig:combined_collages}
\end{figure}

To better understand the qualitative performance of the WARP-CNN, we compared its direct output to the output obtained after a single iteration of the LCA refinement process. This comparison is visualized in \autoref{fig:combined_collages}, where Figure~\ref{fig:collage1} displays the direct reconstructions from the WARP-CNN and Figure~\ref{fig:collage2} shows the reconstructions after applying one LCA iteration. The datasets evaluated include CIFAR-10, STL-10, and Tiny ImageNet.

The direct output of the WARP-CNN appears notably blurry and exhibits a grayish discoloration across all datasets. Additionally, we observed black artifacts in some reconstructed images, particularly evident in the STL-10 reconstructions. For example, in Figure~\ref{fig:collage1}, the reconstructions of Image 2 and Image 3 from the STL-10 dataset show visible dark patches and degraded color fidelity.

{
We hypothesize that these artifacts arise primarily due to the nature of the training cost function and scaling of the targets, which may inadvertently encourage weights that excessively suppress feature activations, resulting in muted reconstructions. With our chosen value for $\gamma = 3$ we place a higher emphasis on smaller targets than larger ones.
}

Interestingly, applying just a single LCA iteration significantly improves the reconstruction quality, as shown in Figure~\ref{fig:collage2}. The discoloration and artifacts are largely eliminated, and the overall image appears cleaner and more consistent with the original input. Although the reconstruction remains somewhat blurry and lacks certain high-frequency details, the one-step LCA refinement produces images that are noticeably sharper and structurally more accurate. This demonstrates that even minimal iterative refinement can substantially enhance WARP-CNN outputs by correcting initial distortions and reducing artifacts.

\subsection{Denoising Performance in Classification Pipelines}
To evaluate the robustness of our WARP-LCA model in enhancing the denoising capabilities within classification pipelines, we employed three distinct pre-trained convolutional neural network architectures: DenseNet-40 with a growth rate of 12 on CIFAR-10 \citep{huang2017densely}, Wide Residual Network with depth 40 and widening factor 8 on CIFAR-10 \citep{zagoruyko2016wide}, and ResNeXt with 29 layers and cardinality 32 using 4D expansions \citep{xie2017aggregated}.

The test set images from CIFAR-10 were subjected to varying levels of additive Gaussian noise. For each noise level and each backbone architecture, two denoising strategies were applied: traditional LCA-based denoising, run for 800 iterations, and our WARP-LCA-based denoising, run for 200 iterations. This approach allowed us to directly compare the efficacy of our WARP-LCA model against the standard LCA under identical noise conditions, consistent with earlier experiments demonstrating comparable PSNR values between the models. Although different levels of noise typically require adjustments to the sparsity parameter $\lambda$, we opted for a $\lambda$ of 0.2 across all noise levels, drawing from methodologies similar to those reported by \citet{teti2022lcanets}. Denoising performance was quantitatively assessed by measuring the classification accuracy post-denoising. 

\begin{figure}[t!] 
    \centering
    \includegraphics[width=\textwidth]{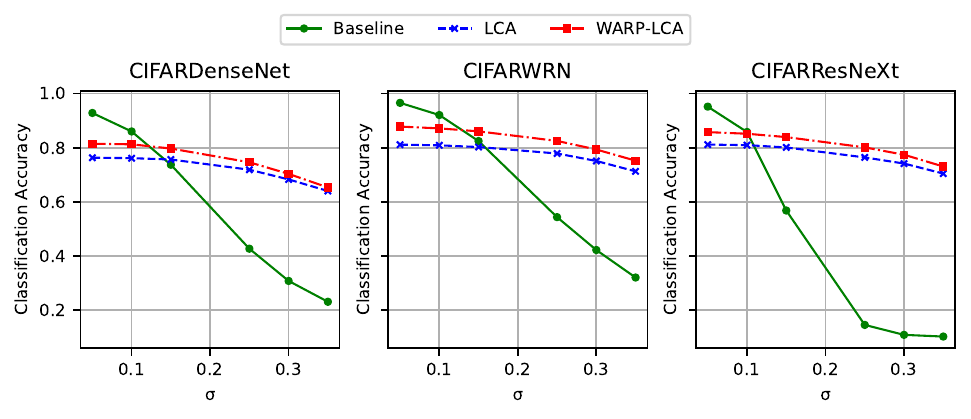}
        \caption{
        \textbf{Testing Robustness against Noise} 
        The graph shows evaluations of CIFAR-10 on DenseNet, WideResenet and ResNext with WARP-LCA and LCA frontend for varying levels of additive gaussian noise ($\sigma$ is the standard deviation).
        }
    \label{fig:classification_accuracy}
\end{figure}

As shown in \autoref{fig:classification_accuracy}, WARP-LCA consistently outperformed traditional LCA in terms of classification accuracy across all tested noise levels and models.

\subsection{Testing WARP-LCA on Larger Images}

To evaluate the scalability of WARP-LCA, we selected a smaller model with lower training and validation performance than the original WARP-CNN described in Section~3 (for model details, see the appendix). This choice was driven by GPU memory constraints when processing larger images during evaluation. The test image, a natural scene obtained from Pexels, has dimensions of $1\,216 \times 1\,824$ pixels.

\begin{figure}[t!]
    \centering
    \begin{subfigure}[b]{\linewidth}
        \centering
        \includegraphics[width=\linewidth]{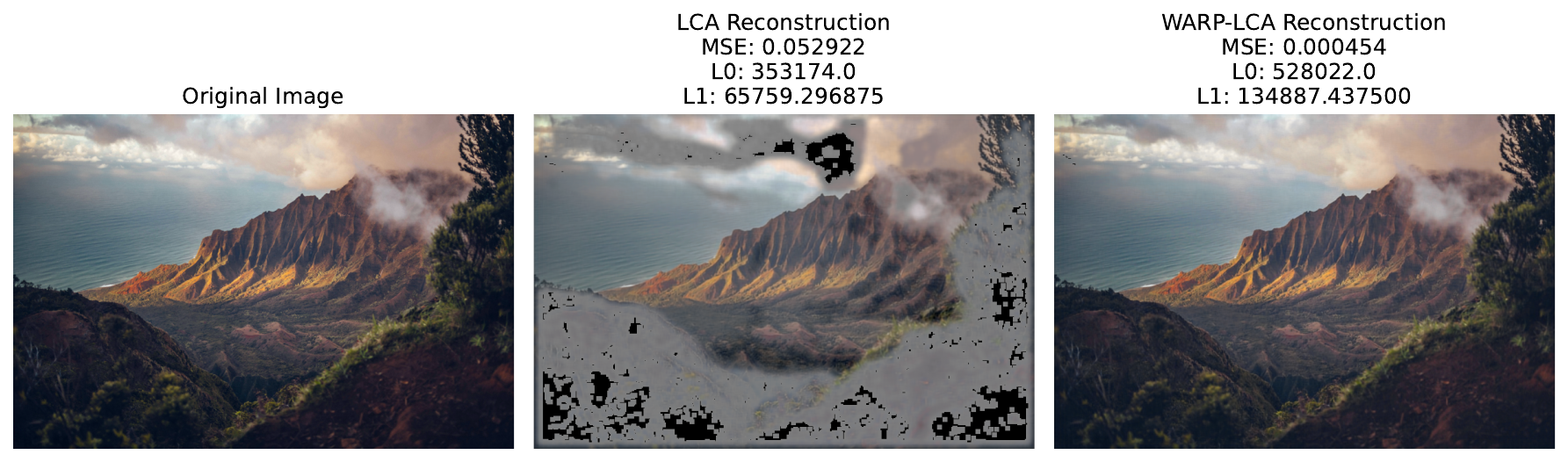}
        \caption{100 iterations with \(\lambda = 0.15\) and \(\tau = 200\)}
        \label{fig:natural_scene_row1}
    \end{subfigure}
    
    \vskip\baselineskip  
    \begin{subfigure}[b]{\linewidth}
        \centering
        \includegraphics[width=\linewidth]{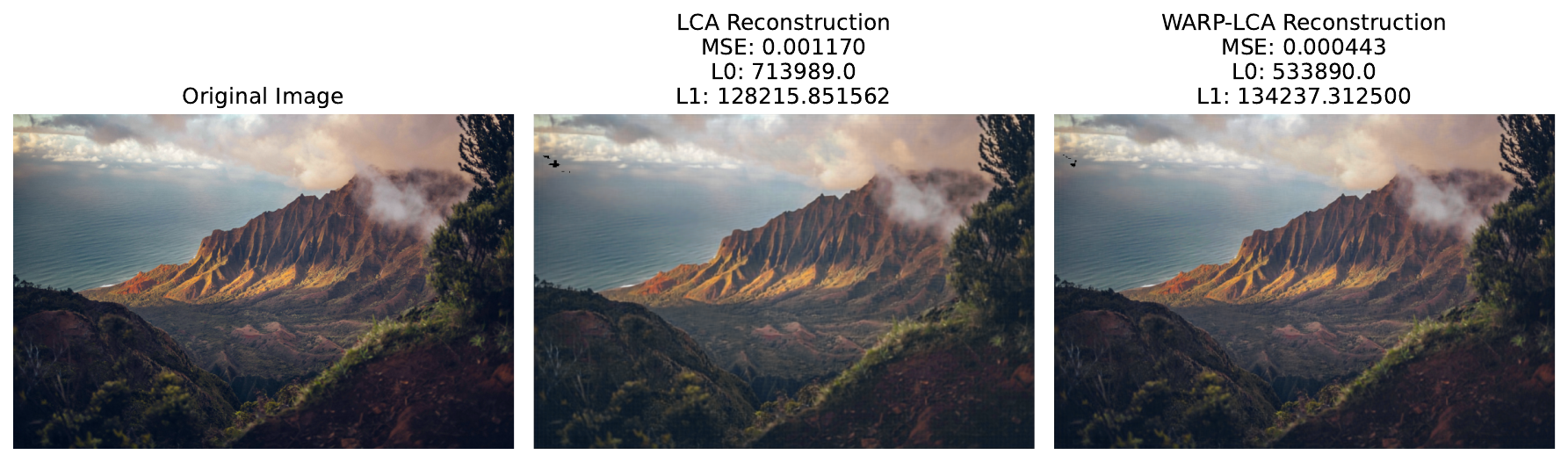}
        \caption{100 iterations with \(\lambda = 0.15\) and \(\tau = 400\)}
        \label{fig:natural_scene_row2}
    \end{subfigure}
    
    \caption{Comparison of WARP-LCA and LCA reconstructions on a natural scene \citep{pexels_image}. The left column shows the original image, the middle image displays the result of WARP-LCA and the right image the result of LCA after 100 iterations with \(\lambda = 0.15\), \(\tau = 200\) and 400 respectively.}
    
    \label{fig:natural_scene_combined}
\end{figure}

\begin{figure}[t!]
    \centering
    \includegraphics[width=\textwidth]{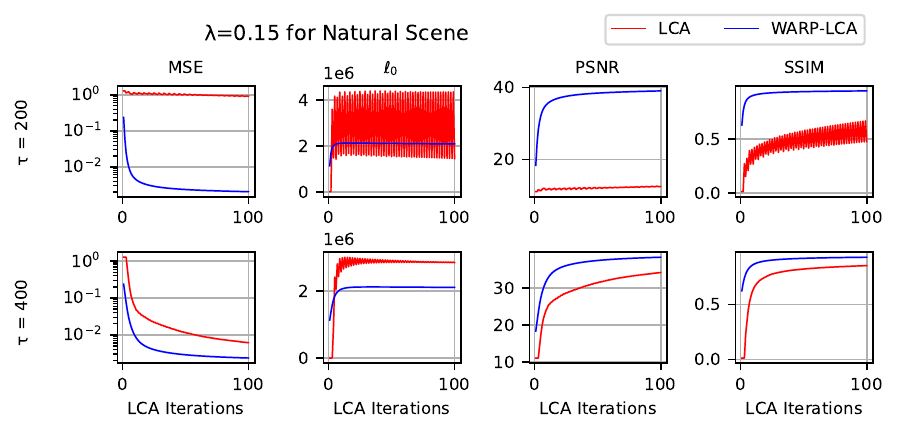}
        \caption{
    \textbf{Encoding Performance Comparison between WARP-LCA and LCA at $\lambda = 0.15$.}
    The figure displays the performance metrics for WARP-LCA and LCA. 
    The columns represent different metrics: Log Mean Squared Error (MSE), \( \ell_0 \)-Norm, Peak Signal-to-Noise Ratio (PSNR), and Structural Similarity Index (SSIM). 
    For each plot, WARP-LCA and LCA Mean and Standard Deviation of the Metrics are shown for comparison.} 
    \label{fig:compare_encoding}
\end{figure}

For this experiment, we used a WARP-LCA model pretrained on the CIFAR-10 dataset, alongside a dictionary similarly trained on CIFAR-10 for comparison with LCA. The results, as depicted in \autoref{fig:natural_scene_combined} and \autoref{fig:compare_encoding}, show a dramatic improvement in convergence speed. All tracked metrics converged even faster than observed on the CIFAR-10, STL10 or Tiny Imagenet test sets.

As shown in \autoref{fig:natural_scene_row1} and the first row of \autoref{fig:compare_encoding} , increasing the learning rate (i.e., decreasing \(\tau\)) causes standard LCA to converge very slowly and unstably. In contrast, WARP-LCA remains stable and continues to converge rapidly, even with the increased step size.
This behavior was consistently observed across experiments on CIFAR-10, STL-10, and Tiny ImageNet, where WARP-LCA effectively handled lower learning rates, while standard LCA using the same dictionary struggled to converge. We attribute this robustness to the high-quality initialization provided by the WARP-CNN, which enables WARP-LCA to tolerate larger step sizes and achieve even faster convergence.

The fully convolutional nature of the trained WARP-CNN facilitates pretraining encoders on smaller datasets and subsequently applying them to much larger images. In such cases, the benefits of using a warm start approach become more pronounced; specifically, the larger the image, the greater the advantage gained from initializing with pretrained encoders. This scalability underscores the effectiveness of WARP-LCA in handling high-resolution, natural images while maintaining efficient convergence behavior.

\subsection{WARP-LCA beyond hard thresholding}
In this section, we investigate how WARP-LCA behaves when replacing the standard hard thresholding (associated with the \(\ell_0\) norm) by various alternatives. Specifically, we compare soft thresholding (\(\ell_1\)), hard thresholding (\(\ell_0\)), an \(\ell_{1/2}\) approach, and a continuous exact \(\ell_0\) (CEL0) method on image reconstruction tasks for CIFAR, STL-10, and Tiny ImageNet. The different thresholding and proximal operators were chosen to be the same as in the work of \citet{calatroni2023beyond} (Beyond \(\ell_1\) Sparse Coding in V1), as their formulations integrate well with the LCA framework. Our primary goal is to analyze reconstruction fidelity and sparsity under each thresholding rule, with metrics including mean squared error (MSE), \(\ell_0\)-norm (approximated by the number of nonzeros), peak signal-to-noise ratio (PSNR), and structural similarity index measure (SSIM). The parameters were set to \(\tau = 200\) and the algorithm was executed for 100 iterations.

First, the standard hard thresholding (proximal operator for \(\ell_0\)) retains any coefficient \(x\) whose magnitude exceeds a threshold \(\tau\), while zeroing out smaller coefficients:
\[
    \mathcal{T}_{\text{hard}}(x; \tau) = 
    \begin{cases}
        x, & \lvert x \rvert \ge \tau,\\[6pt]
        0, & \lvert x \rvert < \tau.
    \end{cases}
\]
Next, the soft thresholding operator (\(\ell_1\)) shrinks coefficients by \(\tau\) in magnitude and sets to zero those below the threshold:
\[
    \mathcal{T}_{\text{soft}}(x; \tau) = 
    \mathrm{sign}(x)\,\max\bigl(\lvert x\rvert - \tau,\;0\bigr).
\]
We also employ an \(\ell_{1/2}\) thresholding operator as shown in \citet{xu2012l_}, which is more aggressive in promoting sparsity than \(\ell_1\) proximal operator. A closed-form solution does not appear as straightforward as in the hard or soft cases, yet it can be expressed through the following piecewise definition:
\[
    \Xi_{\theta,1/2}(r^*) = 
    \begin{cases}
      f_{\theta,1/2}(r^*), & \lvert r^* \rvert > \sqrt[3]{54}\,\theta^{2/3},\\[6pt]
      0, & \text{otherwise},
    \end{cases}
\]
where
\[
    f_{\theta,1/2}(r^*) \;=\; \frac{2\,r^*}{3}\,\Bigl(1 + \cos\Bigl(\tfrac{2\pi}{3} \;-\; \tfrac{2}{3}\,\psi_{\theta}(r^*)\Bigr)\Bigr),
\]
and
\[
    \psi_{\theta}(r^*) 
    \;=\; \arccos\!\Bigl(\,\tfrac{\theta}{8}\,\Bigl(\tfrac{\lvert r^*\rvert}{3}\Bigr)^{-\tfrac{3}{2}}\Bigr).
\]
In practice, the \(\ell_{1/2}\) thresholding operator solution as delineated in \citep{liang2023reduced}, which presents an algorithm yielding a \(k\)-sparse solution, is utilized. This approach employs an alternative sparsity criterion compared to selecting a regularization parameter \(\lambda\); however, for the purpose of contrasting a “warped” variant with the standard formulation, this methodological difference suffices for comparative analysis.

Finally, we consider the CEL0 approach (\citet{soubies2015continuous}), which serves as a continuous relaxation of \(\ell_0\) and leads to a thresholding operator that switches between a soft-like behavior and a strict cut-off depending on the norm of the column \(\phi_i\). Denoting \(\lambda\) as a regularization parameter, \(\mu\) as an additional scaling factor, and \(\|\phi_i\|\) as the \(\ell_2\)-norm of the \(i\)-th dictionary column, the CEL0 threshold can be written as

\[
\resizebox{\linewidth}{!}{$
    \Theta^{\text{CEL0}}_{\mu,\lambda}\bigl(r_i^*\bigr) = 
    \begin{cases}
      \displaystyle
      \mathrm{sign}(r_i^*)\,\min\Bigl\{\,\lvert r_i^*\rvert,\;\Bigl(\lvert r_i^*\rvert - \sqrt{2\lambda}\,\|\phi_i\|\Bigr)_+ \Big/\Bigl(1 \;-\; \tfrac{1}{\|\phi_i\|^2}\,\mu \Bigr)\Bigr\},
      & \|\phi_i\|^2\,\mu < 1,\\[12pt]
      r_i^*\;\mathbf{1}_{\{\lvert r_i^*\rvert > \sqrt{2\mu\lambda}\,\|\phi_i\|\}}
      \;+\;\{0,\,r_i^*\}_{\{\lvert r_i^*\rvert = \sqrt{2\mu\lambda}\,\|\phi_i\|\}},
      & \|\phi_i\|^2\,\mu \ge 1.
    \end{cases}
$}
\]

Here, \((\,\cdot\,)_+\) denotes the positive part, i.e., \((z)_+ = \max\{0,\,z\}\). This thresholding rule smoothly interpolates between a near-hard thresholding regime and a softening behavior, depending on \(\mu\) and the local dictionary norm \(\|\phi_i\|\).

Having established how each thresholding operator is defined and integrated into the LCA refinement process, we now turn to a thorough experimental comparison on the specified datasets, assessing reconstruction quality and sparsity.

{
Table~\ref{tab:final_metrics_summary} summarizes performance comparisons between WARP-LCA and standard LCA across CIFAR10, STL10, and TinyImg datasets under different sparsity constraints ($L_0$, $L_1$, $L_{\frac{1}{2}}$, and CE$L_0$). WARP-LCA consistently achieves better reconstruction metrics compared to LCA.

For CIFAR10, WARP-LCA particularly excels under the $L_{\frac{1}{2}}$ constraint, significantly outperforming LCA by dramatically reducing MSE and substantially improving PSNR and SSIM.

On STL10, WARP-LCA exhibits remarkable superiority under the $L_0$ constraint, achieving notably higher PSNR, lower MSE, and enhanced SSIM relative to standard LCA.

For TinyImg, WARP-LCA notably surpasses LCA under the CE$L_0$ constraint, clearly improving PSNR, SSIM, and reducing MSE.

WARP-LCA consistently and significantly improves reconstruction quality across datasets and constraints, with the most pronounced gains observed in the $L_{\frac{1}{2}}$ and $L_0$ scenarios. There were almost no instances where standard LCA outperformed WARP-LCA, except for a negligible difference under the $L_{\frac{1}{2}}$ constraint for TinyImg, where LCA showed slightly lower sparsity ($L_0$ norm) despite inferior reconstruction quality.
}

\begin{table}[H]
\centering
\scriptsize
\caption{Final Metrics Summary Across Datasets and Algorithms}
\begin{tabularx}{\linewidth}{lRRRR}
\toprule
Algorithm & MSE & $L_0$ Norm & PSNR & SSIM \\
\midrule
\multicolumn{5}{l}{\textbf{CIFAR10}} \\
\midrule
$L_0$ WARP-LCA & \bf{0.0112} & \bf{1356.51} & \bf{31.15} & \bf{0.96} \\
$L_0$ LCA      & 0.0251      & 1559.46 & 27.62    & 0.91    \\
\specialrule{0.1pt}{1.5pt}{1.5pt}
$L_1$ WARP-LCA & \bf{0.0157}  & \bf{2209.44} & \bf{29.64}    & \bf{0.94}    \\
$L_1$ LCA      & 0.0312      & 3465.04 & 26.67    & 0.90    \\
\specialrule{0.1pt}{1.5pt}{1.5pt}
$L_{\frac{1}{2}}$ WARP-LCA & \bf{0.0168} & \bf{1161.61} & \bf{29.28} & \bf{0.93} \\
$L_{\frac{1}{2}}$ LCA      & 0.2914 & 2336.29    & 19.48    & 0.71    \\
\specialrule{0.1pt}{1.5pt}{1.5pt}
CE$L_0$ WARP-LCA & \bf{0.0156} & \bf{1481.46}    & \bf{29.70}    & \bf{0.94}    \\ 
CE$L_0$ LCA      & 0.0363 & 1853.08    & 26.47    & 0.89    \\
\midrule
\multicolumn{5}{l}{\textbf{STL10}} \\
\midrule
$L_0$ WARP-LCA & \bf{0.0118} & \bf{10466.21} & \bf{31.83} & \bf{0.94} \\
$L_0$ LCA      & 0.0923      & 12281.02 & 24.91    & 0.83    \\
\specialrule{0.1pt}{1.5pt}{1.5pt}
$L_1$ WARP-LCA & \bf{0.0152}     & \bf{18532.00} & \bf{30.71}    & \bf{0.92}    \\
$L_1$ LCA      & 0.0277      & 31702.58 & 28.07    & 0.88    \\
\specialrule{0.1pt}{1.5pt}{1.5pt}
$L_{\frac{1}{2}}$ WARP-LCA & \bf{0.1812} & 1713.08  & \bf{20.00}    & \bf{0.62}    \\
$L_{\frac{1}{2}}$ LCA      & 0.5954 & \bf{1633.49} & 15.19 & 0.43    \\
\specialrule{0.1pt}{1.5pt}{1.5pt}
CE$L_0$ WARP-LCA & \bf{0.0164} & \bf{11864.79} & \bf{30.30}    & \bf{0.92}    \\
CE$L_0$ LCA      & 0.2431 & 13629.78 & 19.98    & 0.72    \\
\midrule
\multicolumn{5}{l}{\textbf{TinyImg}} \\
\midrule
$L_0$ WARP-LCA & \bf{0.0398} & \bf{6680.87}  & \bf{27.07} & \bf{0.90} \\
$L_0$ LCA      & 0.0992      & 6686.10  & 23.45    & 0.80    \\
\specialrule{0.1pt}{1.5pt}{1.5pt}
$L_1$ WARP-LCA & \bf{0.0491}      & \bf{10006.09} & \bf{26.13}    & \bf{0.88}    \\
$L_1$ LCA      & 0.0839      & 15048.11 & 23.79    & 0.80    \\
\specialrule{0.1pt}{1.5pt}{1.5pt}
$L_{\frac{1}{2}}$ WARP-LCA & \bf{0.1833} & 861.47   & \bf{19.89}    & \bf{0.61}    \\
$L_{\frac{1}{2}}$ LCA      & 0.4952 & \bf{808.62}  & 16.08    & 0.46    \\
\specialrule{0.1pt}{1.5pt}{1.5pt}
CE$L_0$ WARP-LCA & \bf{0.0511} & \bf{7045.30}  & \bf{25.96}    & \bf{0.88}    \\
CE$L_0$ LCA      & 0.1863 & 7517.97  & 20.78    & 0.73    \\
\midrule
\bottomrule
\end{tabularx}
\label{tab:final_metrics_summary}
\end{table}

\section{Discussion}
The comparative analysis in \autoref{sec:experiments} (illustrated in \autoref{fig:compare_multiple_metrics}) compellingly demonstrates the superior efficacy of WARP-LCA, over the traditional LCA: faster convergence and settling for deeper minima. This attribute was consistently evident across a range of noise levels and architectural backbones in the denoising experiment in Section~4.1. These results not only corroborate but also extend the findings by \citet{teti2022lcanets}, propelling WARP-LCA beyond the current state-of-the-art. This advancement underscores the potential of predictive initializations to redefine performance benchmarks in the field, offering robust, fast, and more accurate sparse coding.

The robustness of WARP-LCA, particularly for out-of-distribution data, was tested on the Oxford Pets dataset \citep{parkhi2012cats}, which starkly contrasts with CIFAR-10 in terms of image size (images in the Oxford Pets datset are much larger) and image statistics. The Oxford Pets dataset is a collection of images featuring 37 different breeds of cats and dogs, each annotated with class labels, bounding boxes, and pixel-level segmentation masks. Despite these considerable differences, and the fact that our models were initially trained only on CIFAR images, the fully convolutional nature of predictor architectures allowed for an efficient adaptation to this new context. 

\begin{figure}[t!]
    \centering
    \includegraphics[width=\textwidth]{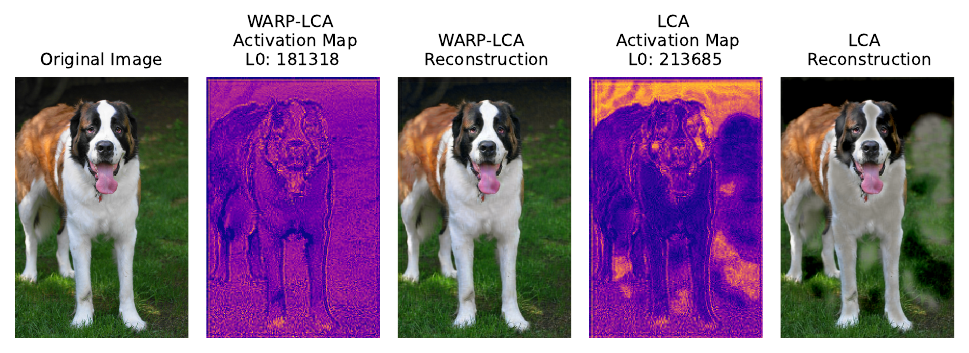}
    \caption{
    \textbf{Accumulated activation maps for WARP-LCA and LCA.}
    This figure presents the original image and the activation maps generated by WARP-LCA and LCA, with a sparsity level ($\lambda = 0.15$). 
    Accumulated activation maps are computed as the count of non-zero sparse coefficients across different channels  (normalized between 0 and 1 across both activation maps to highlight relative activation levels). 
    The WARP-LCA map is derived after 10 iterations, whereas the LCA map results from 300 iterations.
    }
    \label{fig:activation_maps}
\end{figure}

The (relative) accumulated activation map generated by WARP-LCA provides more homogeneous and less extreme structures than the accumulated LCA activation map, indicated much more uniformly distributed activations. Even after 300 iterations, activity in the accumulated LCA map is still concentrated in very dark and light areas of the image. In addition, WARP-LCA also has a lower \( \ell_0 \)-Norm than the LCA. The reconstructed image from the LCA appears not to have converged yet, as the colors, especially in the light and dark areas, seems to be faded. The reconstructed image from the WARP-LCA seems to be much closer to the original.

This may imply that the CIFAR dictionary alone may be sufficient to be used on data other than CIFAR images and WARP-LCA managed to predict states that lead to a consistent and balanced activation profile.

We have experimented with different approaches to warm starting LCA, including random initialization but found that the only methods that lead to convergence are the conventional zero initialization as is typical for LCA and our CNN predictor. We have compared the similarity of learned cifar10-kernels to the correlation of activations of LCA and WARP-LCA after 1000 optimization cycles up to convergence and found that the WARP-LCA method has lower correlation between kernels as would be expected from the earlier results. The prediction of initial states leads to more diverse and robust minima (please see the appendix for a comparrison of LCA, WARP-LCA, CNN-only prediction and simple convolution).

During our exploration, we opted for a balance between model complexity and effectiveness, as detailed in \autoref{tab:branchnet_forward}. While we assessed performance at various sparsity levels, notably at lambda=0.85, the predictive performance of WARP-LCA diminished, as documented in the appendix. This reduction likely stems from the restricted solution space at higher sparsity levels, where fewer unique kernel combinations are viable for reconstructing the input. In contrast, at lower lambdas, it was possible to achieve minima that significantly outperformed traditional LCA.

It is important to note that our experiments were conducted using a single, learned over-complete dictionary. The application of WARP-LCA to dictionaries with lower overcompleteness---and hence fewer kernels---might prove more challenging. Yet, a larger dictionary typically enhances the benefits of WARP-CNN in terms of convergence and solution quality, as demonstrated in \autoref{fig:activation_maps}. Unlike the approach by \citep{teti2022lcanets}, we did not train the backbones on sparse activations but rather opted to reconstruct the image, which allowed us to employ pretrained backbones and evaluate WARP-LCA and LCA as distinct preprocessing modules. Training a classification model on sparse activations obtain by WARP-LCA may provide more robustness.

It should be noted as well that neither LCA nor WARP-LCA can reconstruct or fill missing areas beyond local disturbances with semantically useful information, since the principle does not provide top-down, that is, high-level and context-depedent projections, which is accomplished e.g. with generative models.

\subsection{Limitations}
While our results underscore the potential of WARP-LCA, several limitations remain. In WARP-LCA, a convolutional neural network (CNN) first provides a warm start for the sparse coefficients, after which an iterative LCA procedure refines those coefficients. This two-stage process naturally incurs additional computational cost relative to running LCA alone or a standalone CNN. Specifically, the input must be passed through the CNN once, and then multiple LCA iterations are performed, each updating the sparse coefficients. The computational overhead becomes more pronounced as input sizes grow or the number of refinement steps increases.

As a concrete example, in Section~4.3 we had to reduce the size of the trained CNN for larger images so that it would fit in GPU memory; using the full architecture would have exceeded our hardware resources. We did not exhaustively explore potential network architectures or perform a broad hyperparameter grid search to identify more parameter-efficient (or memory-friendly) configurations. Consequently, the scalability of WARP-LCA—especially in terms of network depth, input resolution, and number of iterative updates warrants deeper investigation.

In addition, future efforts could extend the WARP-LCA framework to deep or hierarchical sparse coding models \citep{boutin2021sparse}, including deep convolutional variants, thereby enabling richer feature hierarchies and more expressive representations. We also have not explored the full range of contemporary LCA applications, such as robust model inversion attacks \citep{dibbo2025improving}, on-chip learning \citep{takaghaj2024exemplar}, or robust audio classification \citep{dibbo2024lcanets}. Given that WARP-LCA demonstrates improvements across multiple algorithms and datasets, it is very likely these advantages would also translate effectively into such domains. Lastly, further work on adaptive models that not only estimate the initial states but also automatically learn hyperparameters—such as the optimal number of iterations and the sparsity penalty parameter \(\lambda\)—could help reduce manual tuning, improving both efficiency and practicality. These advancements would help bridge the gap between the current theoretical framework and large-scale deployment scenarios.

\section{Conclusion}
This research introduces WARP-LCA, an enhanced version of the Locally Competitive Algorithm (LCA), and demonstrates its efficacy in terms of accelerating convergence, improving robustness to noise and the initial choice of parameters, and enabling the algorithm to achieve deeper minima and hence better solutions. As compared to the original LCA, we consistently observed improvements of image reconstruction metrics across various noise levels, and improvements of noise robustness across several deep neural network backbone architectures. 

Our investigations further revealed the robust generalizability of WARP-LCA, particularly with the Oxford Pets dataset, which significantly differed from the training dataset (CIFAR) in image size, type, and image statistics. Moreover, WARP-LCA proved highly effective in initializing other thresholding algorithms, such as CELO and L1/2, leading to substantially improved outcomes. By providing a strong initial estimate that aligns well with the underlying data structure, WARP-LCA facilitates faster convergence and enhances the precision of these subsequent methods, yielding significantly better results in practice. Additionally, we found that a good initialization—such as that provided by WARP-CNN—permits the use of a larger step size during optimization, thereby enabling even faster convergence while maintaining stability and accuracy.

Looking ahead, expanding the capabilities of WARP-LCA to directly process noisy inputs could dramatically enhance its applicability for denoising, robustness, and other sparse coding applications. Furthermore, alternative approaches—such as specialized application-tailored models or deep sparse coding architectures—could be employed to complement or extend the current framework, offering additional pathways for improved performance and flexibility in diverse scenarios.

\newpage
\bibliographystyle{elsarticle-harv}
\bibliography{cas-refs}

\newpage
\section{Appendix}
\label{sec:appendix}

\subsection{Parameters for Training LCA on CIFAR, STL10 and Tiny Imagenet}
\scriptsize
\begin{table}[H]
\caption{LCA hyperparameters on CIFAR-10.}
\label{tab:lca_hyperparameters}
\centering
\begin{tabular}{ l c }
\toprule
Hyperparameter & Value \\
\midrule
Output neurons & 100 \\
Input neurons & 3 \\
Kernel size & 9 \\
Stride & 2 \\
Lambda & 2.55 \\
Tau & 100 \\
Eta & 0.01 \\
Lca iters & 800 \\
Pad & same \\
Nonneg & True \\
Transfer func & soft\_threshold \\
\bottomrule
\end{tabular}
\end{table}

\newpage
\subsection{WARP-CNN Architecture and Training hyper-parameters}
\begin{table}[h!]
    \centering
    \caption{WARP-CNN Architecture}
    \label{tab:branchnet}
    \begin{adjustbox}{max width=\textwidth}
    \begin{tabular}{lllc}
        \toprule
        Layer & Type & Parameters & Output Shape \\
        \midrule
        \textbf{Common Layers} & & & \\
        Conv1 & Conv2d & 4, 512, kernel\_size=5, stride=1, padding=2 & (batch\_size, 512, H, W) \\
        BN1 & BatchNorm2d & 512 & (batch\_size, 512, H, W) \\
        Conv2 & Conv2d & 512, 512, kernel\_size=5, stride=1, padding=2 & (batch\_size, 512, H, W) \\
        BN2 & BatchNorm2d & 512 & (batch\_size, 512, H, W) \\
        Conv3 & Conv2d & 512, 512, kernel\_size=3, stride=1, padding=1 & (batch\_size, 512, H, W) \\
        BN3 & BatchNorm2d & 512 & (batch\_size, 512, H, W) \\
        Conv4 & Conv2d & 512, 512, kernel\_size=3, stride=1, padding=1 & (batch\_size, 512, H, W) \\
        BN4 & BatchNorm2d & 512 & (batch\_size, 512, H, W) \\
        Dropout & Dropout & p=0.3 & (batch\_size, 512, H, W) \\
        Conv5 & Conv2d & 512, 512, kernel\_size=3, stride=1, padding=1 & (batch\_size, 512, H, W) \\
        BN5 & BatchNorm2d & 512 & (batch\_size, 512, H, W) \\
        Conv6 & Conv2d & 512, 512, kernel\_size=3, stride=1, padding=1 & (batch\_size, 512, H, W) \\
        BN6 & BatchNorm2d & 512 & (batch\_size, 512, H, W) \\
        AdjustChannels & Conv2d & 512, 256, kernel\_size=1 & (batch\_size, 256, H, W) \\
        \midrule
        \textbf{Downward Branch} & & & \\
        Conv\_d1 & Conv2d & 512, 256, kernel\_size=3, stride=1, padding=1 & (batch\_size, 256, H, W) \\
        BN\_d1 & BatchNorm2d & 256 & (batch\_size, 256, H, W) \\
        Conv\_d2 & Conv2d & 256, 256, kernel\_size=3, stride=1, padding=1 & (batch\_size, 256, H, W) \\
        BN\_d2 & BatchNorm2d & 256 & (batch\_size, 256, H, W) \\
        Conv\_d3 & Conv2d & 256, 100, kernel\_size=3, stride=2, padding=1 & (batch\_size, 100, H/2, W/2) \\
        BN\_d3 & BatchNorm2d & 100 & (batch\_size, 100, H/2, W/2) \\
        \midrule
        \textbf{Upward Branch} & & & \\
        Conv\_u1 & Conv2d & 512, 256, kernel\_size=3, stride=1, padding=1 & (batch\_size, 256, H, W) \\
        BN\_u1 & BatchNorm2d & 256 & (batch\_size, 256, H, W) \\
        Conv\_u2 & Conv2d & 256, 256, kernel\_size=3, stride=1, padding=1 & (batch\_size, 256, H, W) \\
        BN\_u2 & BatchNorm2d & 256 & (batch\_size, 256, H, W) \\
        Conv\_u3 & Conv2d & 256, 100, kernel\_size=3, stride=2, padding=1 & (batch\_size, 100, H/2, W/2) \\
        BN\_u3 & BatchNorm2d & 100 & (batch\_size, 100, H/2, W/2) \\
        \bottomrule
    \end{tabular}
    \end{adjustbox}
\end{table}

\newpage

\begin{table}[h!]
    \caption{Forward Pass of WARP CNN with Skip Connections}
    \label{tab:branchnet_forward}
    \centering
    \begin{adjustbox}{max width=\textwidth}
    \begin{tabular}{lllc}
        \toprule
        Step & Operation & Input & Output Shape \\
        \midrule
        1 & Conv1 + ReLU & x & (batch\_size, 512, H, W) \\
        2 & BN1 & Output of Step 1 & (batch\_size, 512, H, W) \\
        3 & Conv2 + ReLU & Output of Step 2 & (batch\_size, 512, H, W) \\
        4 & BN2 & Output of Step 3 & (batch\_size, 512, H, W) \\
        5 & Conv3 + ReLU & Output of Step 4 & (batch\_size, 512, H, W) \\
        6 & BN3 & Output of Step 5 & (batch\_size, 512, H, W) \\
        7 & Conv4 + ReLU & Output of Step 6 & (batch\_size, 512, H, W) \\
        8 & BN4 & Output of Step 7 & (batch\_size, 512, H, W) \\
        9 & Dropout & Output of Step 8 & (batch\_size, 512, H, W) \\
        10 & Conv5 + ReLU & Output of Step 9 & (batch\_size, 512, H, W) \\
        11 & BN5 & Output of Step 10 & (batch\_size, 512, H, W) \\
        12 & Dropout & Output of Step 11 & (batch\_size, 512, H, W) \\
        13 & Conv6 + ReLU & Output of Step 12 & (batch\_size, 512, H, W) \\
        14 & BN6 & Output of Step 13 & (batch\_size, 512, H, W) \\
        15 & Dropout & Output of Step 14 & (batch\_size, 512, H, W) \\
        16 & AdjustChannels & Output of Step 15 & (batch\_size, 256, H, W) \\
        \midrule
        \textbf{Downward Branch} & & & \\
        \midrule
        17 & Conv\_d1 + ReLU & Output of Step 15 & (batch\_size, 256, H, W) \\
        18 & BN\_d1 & Output of Step 17 & (batch\_size, 256, H, W) \\
        19 & Conv\_d2 + ReLU & Output of Step 18 + Output of Step 16 & (batch\_size, 256, H, W) \\
        20 & BN\_d2 & Output of Step 19 & (batch\_size, 256, H, W) \\
        21 & Conv\_d3 + ReLU & Output of Step 20 & (batch\_size, FEATURES, H/2, W/2) \\
        22 & BN\_d3 & Output of Step 21 & (batch\_size, FEATURES, H/2, W/2) \\
        \midrule
        \textbf{Upward Branch} & & & \\
        \midrule
        23 & Conv\_u1 + ReLU & Output of Step 15 & (batch\_size, 256, H, W) \\
        24 & BN\_u1 & Output of Step 23 & (batch\_size, 256, H, W) \\
        25 & Conv\_u2 + ReLU & Output of Step 24 + Output of Step 16 & (batch\_size, 256, H, W) \\
        26 & BN\_u2 & Output of Step 25 & (batch\_size, 256, H, W) \\
        27 & Conv\_u3 + ReLU & Output of Step 26 & (batch\_size, FEATURES, H/2, W/2) \\
        28 & BN\_u3 & Output of Step 27 & (batch\_size, FEATURES, H/2, W/2) \\
        \midrule
        29 & Subtraction & Output of Step 22 - Output of Step 28 & (batch\_size, FEATURES, H/2, W/2) \\
        30 & Sigmoid & Output of Step 29 & (batch\_size, FEATURES, H/2, W/2) \\
        \bottomrule
    \end{tabular}
    \end{adjustbox}
\end{table}

\newpage 
\begin{table}[h!]
\centering
\scriptsize
\caption{Architectural specifications of WARP-CNN variants: WarpNet employs 6 convolutional layers at 512 channels with a channel reduction from 512 to 256 pre-branch and complex downward/upward branches (512→256→100). WarpNet\_Small1 uses 6 convolutional layers at 256 channels, adjusting from 256 to 128 and adapting branches to 256→128→100. WarpNet\_Small2 comprises 3 convolutional layers at 128 channels, reducing from 128 to 64 with branches following 128→64→100. WarpNet\_Small3 contains 2 convolutional layers at 64 channels, transitioning from 64 to 32, with branches using 64→32→100. WarpNet\_Small4 features 2 convolutional layers at 32 channels, adjusting from 32 to 16 and branches utilizing 32→16→100.}

\begin{tabular}{lcc}
\toprule
Model & Final Train Loss & Final Val Loss \\
\midrule
\multicolumn{3}{l}{\textbf{cifar}} \\
\midrule
WarpNet & 0.0037 & 0.0036 \\
WarpNet Small1 & 0.0047 & 0.0042 \\
WarpNet Small2 & 0.0063 & 0.0057 \\
WarpNet Small3 & 0.0081 & 0.0074 \\
WarpNet Small4 & 0.0100 & 0.0091 \\
\midrule
\multicolumn{3}{l}{\textbf{stl10}} \\
\midrule
WarpNet & 0.0044 & 0.0045 \\
WarpNet Small1 & 0.0078 & 0.0068 \\
WarpNet Small2 & 0.0106 & 0.0095 \\
WarpNet Small3 & 0.0133 & 0.0120 \\
WarpNet Small4 & 0.0141 & 0.0129 \\
\midrule
\multicolumn{3}{l}{\textbf{tinyimg}} \\
\midrule
WarpNet & 0.0057 & 0.0050 \\
WarpNet Small1 & 0.0073 & 0.0064 \\
WarpNet Small2 & 0.0100 & 0.0090 \\
WarpNet Small3 & 0.0131 & 0.0123 \\
WarpNet Small4 & 0.0150 & 0.0138 \\

\midrule
\bottomrule
\end{tabular}
\label{tab:models_summary}
\end{table}
\newpage

\begin{table}[h!]
\caption{Training and testing details for the CIFAR-10, STL10 and Tiny ImageNet datasets.}
\label{tab:training_testing_details}
\scriptsize
\centering
\begin{tabular}{ p{2.5cm} p{10.5cm} }
\toprule
\textbf{Detail} & \textbf{Description} \\
\midrule
\textbf{Dataset} & CIFAR-10 standard training and test splits \\
\midrule
\textbf{Data Transformation} & \begin{tabular}[c]{@{}l@{}} 
ToTensor() \\ 
Normalize(mean=[0.4914, 0.4822, 0.4465], std=[0.247, 0.243, 0.261]) \\
\end{tabular} \\
\midrule
\textbf{Network Initialization} & \begin{tabular}[c]{@{}l@{}} 
\texttt{net.apply(lambda m: sparse\_init(m,} \\ 
\texttt{sparsity=0.9, std=0.01))}
\end{tabular} \\
\midrule
\textbf{Sparse Initialization Function} & \begin{tabular}[c]{@{}l@{}} 
\texttt{def sparse\_init(m, sparsity=0.9, std=0.01):} \\
\quad \texttt{if isinstance(m, (nn.Linear, nn.Conv2d)):} \\
\quad \quad \texttt{with torch.no\_grad():} \\
\quad \quad \quad \texttt{sparse\_weights = torch.randn(m.weight.size()) * std} \\
\quad \quad \quad \texttt{mask = torch.rand(m.weight.size()) > sparsity} \\
\quad \quad \quad \texttt{sparse\_weights = sparse\_weights * mask.float()} \\
\quad \quad \quad \texttt{m.weight = nn.Parameter(sparse\_weights)} \\
\quad \quad \quad \texttt{if m.bias is not None:} \\
\quad \quad \quad \quad \texttt{m.bias.data.zero\_()}
\end{tabular} \\
\midrule
\textbf{Optimizer} & \begin{tabular}[c]{@{}l@{}} 
Adam \\ 
Learning Rate = 0.0001 \\ 
\texttt{optimizer = torch.optim.Adam(net.parameters(),} \\ 
\texttt{lr=0.0001)}
\end{tabular} \\
\midrule
\textbf{Loss Function} & \begin{tabular}[c]{@{}l@{}} 
\texttt{custom\_weighted\_mse\_loss\_with\_laplace\_scale} \\
\quad with Laplace-like scaling\\
\quad parameters. Laplace scale: 3.0.
\end{tabular} \\
\midrule
\textbf{Scaling of Target Activations} & Target activations are scaled between 0 and 1 during training, and predicted targets are descaled during inference. \\
\midrule
\textbf{Validation Data} & 1\% of the training data was used for validation \\
\bottomrule
\end{tabular}
\end{table}

\newpage
\subsection{Training procedure for WARP-LCA}

\begin{algorithm}[H]
\caption{Training Dictionary, Encoding Data and Training CNN for WARP-LCA}
\begin{algorithmic}[1]

\State \textbf{Input:} Dataset $D$, Number of epochs $E$, Initial $\lambda$, Increase factor $f$, Number of ISTA steps $n$

\State \textbf{Output:} Trained Dictionary $K$, Encoded Dataset $D'$, Trained CNN

\State Initialize dictionary $K$

\For{$epoch = 1$ to $E$}
    \For{each sample in $D$}
        \State Perform $n$ steps of ISTA with L1 sparsity (soft thresholding)
    \EndFor
    \State $\lambda \gets \lambda \times f$ \Comment{Increase sparsity factor}
\EndFor

\State \textbf{Encode dataset:}
\For{each sample in $D$}
    \State Run LCA with hard thresholding
    \State Save activations to new dataset $D'$
\EndFor

\State \textbf{Train CNN:}
\State Initialize CNN
\For{each (input, label) in ($D$, $D'$)}
    \State Train CNN on input to predict corresponding label
\EndFor

\end{algorithmic}
\end{algorithm}

\subsection{More Examples for Accumulated Aggregation Maps}

\begin{figure}[H]
    \centering
    \includegraphics[width=\textwidth]{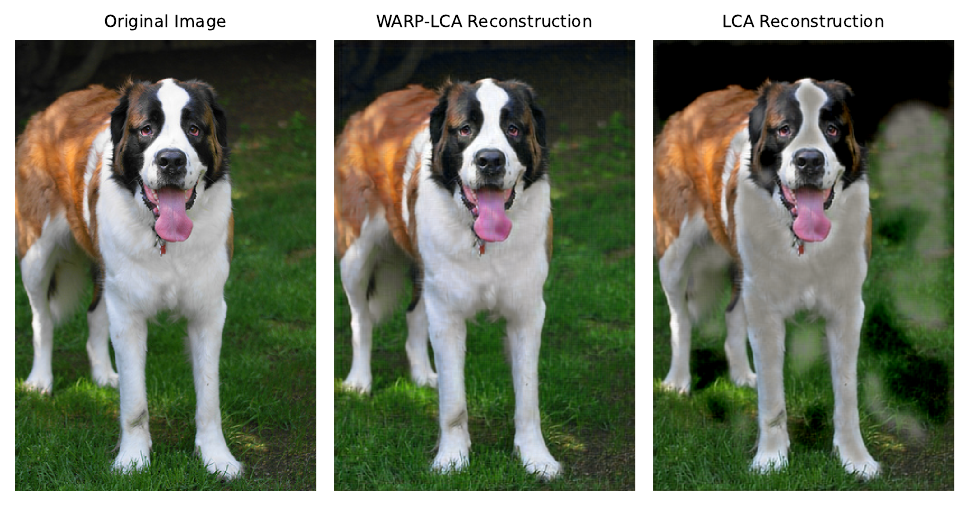}
        \caption{
    \textbf{Larger Example of as shown in \ref{fig:activation_maps}} 
    }
    \label{fig:compare}
\end{figure}

\begin{figure}[H]
    \centering
    \includegraphics[width=\textwidth]{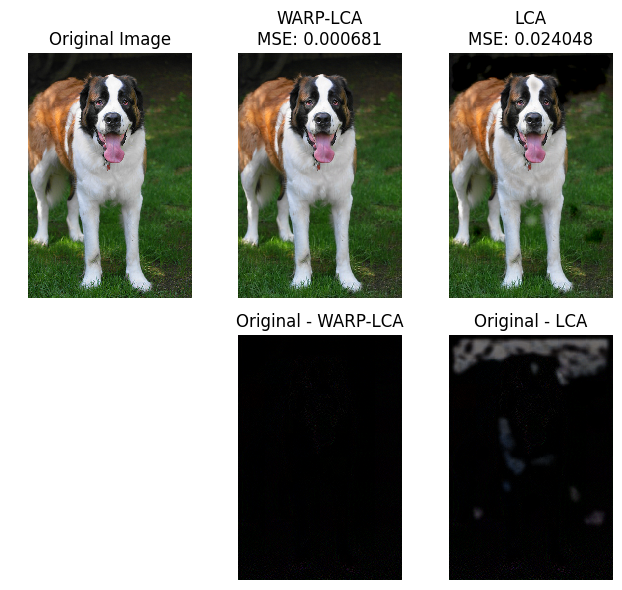}.
        \caption{
    \textbf{Larger Example of as shown in \ref{fig:activation_maps}}. Both WARP-LCA and LCA were executed for 1000 iterations to fully converge. Visibly, the LCA image has a worse reconstruction than the WARP-LCA image. It seems to struggle with very dark or bright areas of the original image as can be seen seen in the difference plot in the second row (subtraction of original image and respective LCA method).
    }
    \label{fig:compare}
\end{figure}

\begin{figure}[H]
    \centering
    \includegraphics[width=\textwidth]{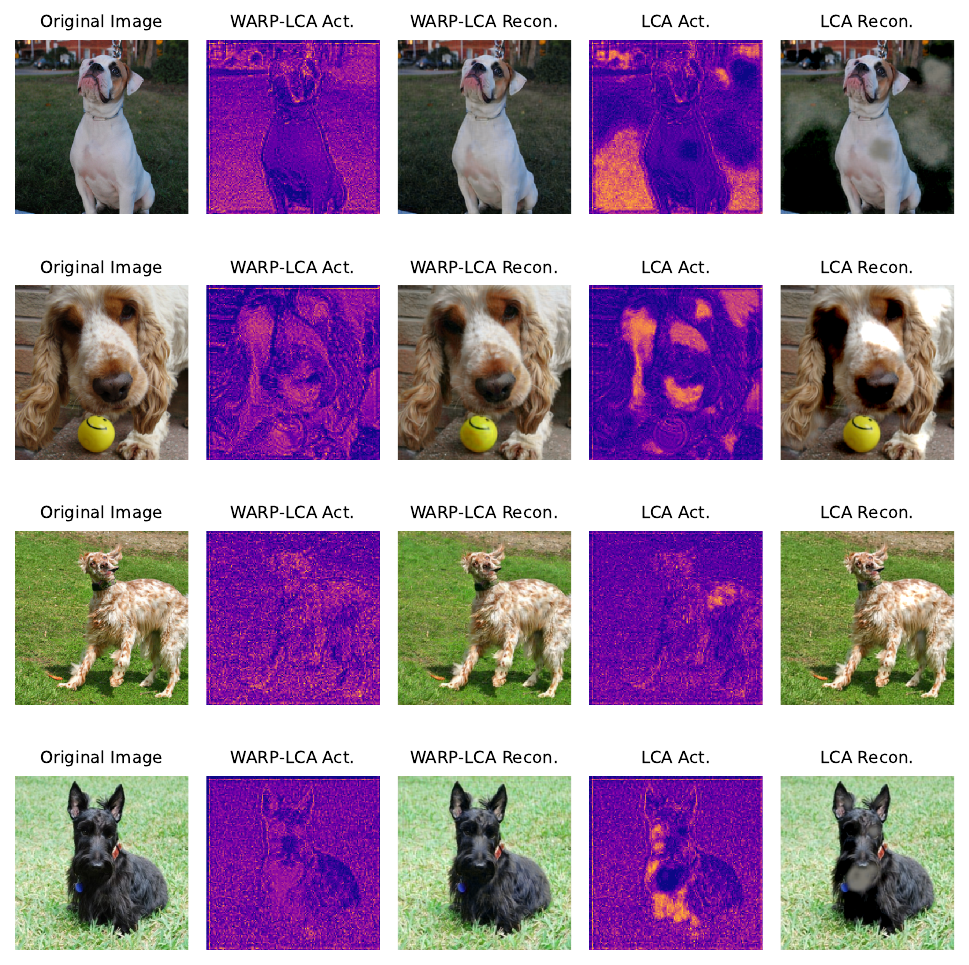}
        \caption{
    \textbf{More Examples for Accumulated Aggregation Maps} 
    }
    \label{fig:compare}
\end{figure}

\end{document}